\documentclass[conference]{IEEEtran}
\IEEEoverridecommandlockouts
\usepackage{cite}
\usepackage{amsmath,amssymb,amsfonts}
\usepackage{algorithmic}
\usepackage{graphicx}
\usepackage{textcomp}
\usepackage{xcolor}
\def\BibTeX{{\rm B\kern-.05em{\sc i\kern-.025em b}\kern-.08em
    T\kern-.1667em\lower.7ex\hbox{E}\kern-.125emX}}

\usepackage{scalerel}
\usepackage{tikz}
\usetikzlibrary{svg.path}

\definecolor{orcidlogocol}{HTML}{A6CE39}
\tikzset{
    orcidlogo/.pic={
        \fill[orcidlogocol] svg{M256,128c0,70.7-57.3,128-128,128C57.3,256,0,198.7,0,128C0,57.3,57.3,0,128,0C198.7,0,256,57.3,256,128z};
        \fill[white] svg{M86.3,186.2H70.9V79.1h15.4v48.4V186.2z}
        svg{M108.9,79.1h41.6c39.6,0,57,28.3,57,53.6c0,27.5-21.5,53.6-56.8,53.6h-41.8V79.1z M124.3,172.4h24.5c34.9,0,42.9-26.5,42.9-39.7c0-21.5-13.7-39.7-43.7-39.7h-23.7V172.4z}
        svg{M88.7,56.8c0,5.5-4.5,10.1-10.1,10.1c-5.6,0-10.1-4.6-10.1-10.1c0-5.6,4.5-10.1,10.1-10.1C84.2,46.7,88.7,51.3,88.7,56.8z};
    }
}

\newcommand\orcidicon[1]{\href{https://orcid.org/#1}{\mbox{\scalerel*{
                \begin{tikzpicture}[yscale=-1,transform shape]
                \pic{orcidlogo};
                \end{tikzpicture}
            }{|}}}}

\usepackage[pagebackref=true,breaklinks=true,colorlinks,bookmarks=false]{hyperref}
\usepackage{balance}
\usepackage{tabularx,booktabs}
\usepackage{array, multirow, boldline}
\usepackage{subcaption}

\begin{document}

\title{Toward Improving Robustness of Object Detectors against Domain Shift}

\author{\IEEEauthorblockN{Le-Anh Tran\IEEEauthorrefmark{2}$^{\textsuperscript{\orcidicon{0000-0002-9380-7166}}}$}
\IEEEauthorblockA{\textit{Dept. of Electronics Engineering} \\
\textit{Myongji University}\\
Yongin, South Korea \\
leanhtran@mju.ac.kr}
\and
\IEEEauthorblockN{Chung Nguyen Tran\IEEEauthorrefmark{2}$^{\textsuperscript{\orcidicon{0009-0000-6402-8799}}}$}
\IEEEauthorblockA{\textit{Dept. of Microelectronics and Electronics Systems} \\
\textit{Universitat Autònoma de Barcelona}\\
Bellaterra, Spain \\
chungnguyen.tran@autonoma.cat}
\and
\IEEEauthorblockN{Dong-Chul Park*$^{\textsuperscript{\orcidicon{0009-0007-7442-2301}}}$}
\IEEEauthorblockA{\textit{Dept. of Electronics Engineering} \\
\textit{Myongji University}\\
Yongin, South Korea \\
parkd@mju.ac.kr}
\and
\IEEEauthorblockN{\quad\quad\quad\quad Jordi Carrabina$^{\textsuperscript{\orcidicon{0000-0002-9540-8759}}}$} 
\IEEEauthorblockA{\quad\quad\quad\quad \textit{Dept. of Microelectronics and Electronics Systems} \\
\textit{\quad\quad\quad\quad Universitat Autònoma de Barcelona}\\
\quad\quad\quad\quad Bellaterra, Spain \\
\quad\quad\quad\quad jordi.carrabina@uab.cat}
\and
\IEEEauthorblockN{David Castells-Rufas$^{\textsuperscript{\orcidicon{0000-0002-7181-9705}}}$}
\IEEEauthorblockA{\textit{Dept. of Microelectronics and Electronics Systems} \\
\textit{Universitat Autònoma de Barcelona}\\
Bellaterra, Spain \\
david.castells@uab.cat}
\thanks{\IEEEauthorrefmark{2}Equal contribution.}
\thanks{*Corresponding author.}
}

\maketitle

\begin{abstract}
This paper proposes a data augmentation method for improving the robustness of driving object detectors against domain shift. Domain shift problem arises when there is a significant change between the distribution of the source data domain used in the training phase and that of the target data domain in the deployment phase. Domain shift is known as one of the most popular reasons resulting in the considerable drop in the performance of deep neural network models. In order to address this problem, one effective approach is to increase the diversity of training data. To this end, we propose a data synthesis module that can be utilized to train more robust and effective object detectors. By adopting YOLOv4 as a base object detector, we have witnessed a remarkable improvement in performance on both the source and target domain data. The code of this work is publicly available at \url{https://github.com/tranleanh/haze-synthesis}.
\end{abstract}



\section{Introduction}

Over the last decade, convolutional neural networks (CNNs) have shown their ubiquitous influence in computer vision with explosive rises in performance on numerous real-world vision tasks such as object detection \cite{bochkovskiy2020yolov4}, image segmentation \cite{tran2019robust}, and image restoration \cite{tran2022novel}. CNN-based models generally are data-driven which can learn to estimate desired outcomes based on the data that they are trained on. This property of CNN, however, is related to one of its principal drawbacks. That is, CNN-based approaches usually overfit the training data while their generalization capability to out-of-domain data is insufficient. In other words, their performance is often degraded when facing \textit{domain shift} which is when there is a dramatic change between the distribution of the source data domain used in the training phase in comparison with that of the target data domain in the deployment phase. Domain shift is extremely difficult to handle in an online fashion with conventional learning methods since CNN-based models must be trained with sufficient time and training data prior to being deployed into any applications.

\begin{figure}[t]
\centering
\includegraphics[width=0.96\columnwidth]{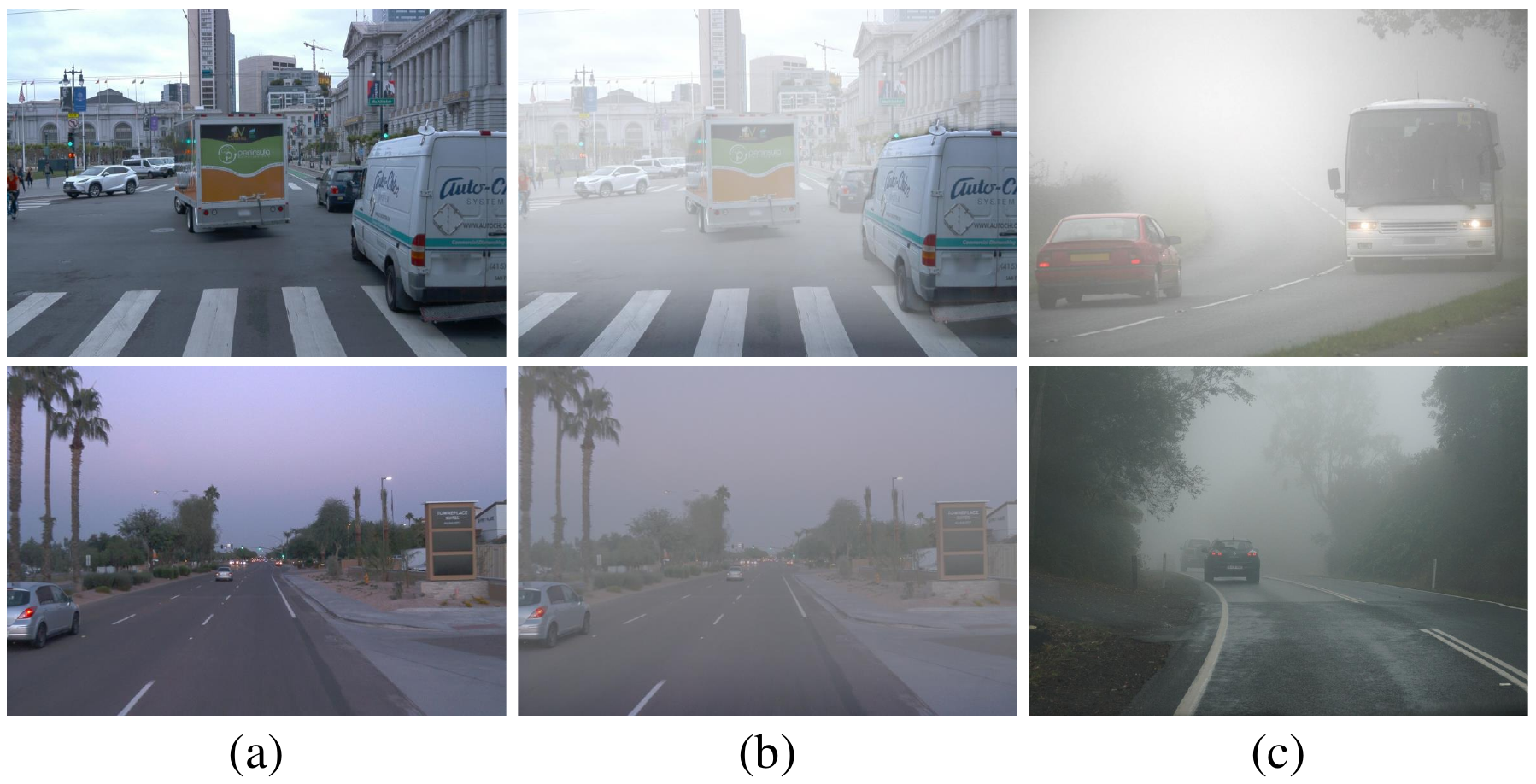}
\caption{Data synthesis: (a) Original clean image, (b) Synthetic image by the proposed method, (c) Natural hazy image \cite{sakaridis2018semantic}.}
\label{fig1:fogexamples}
\end{figure}

\begin{figure*}[t]
\centering
\includegraphics[width=1.0\textwidth]{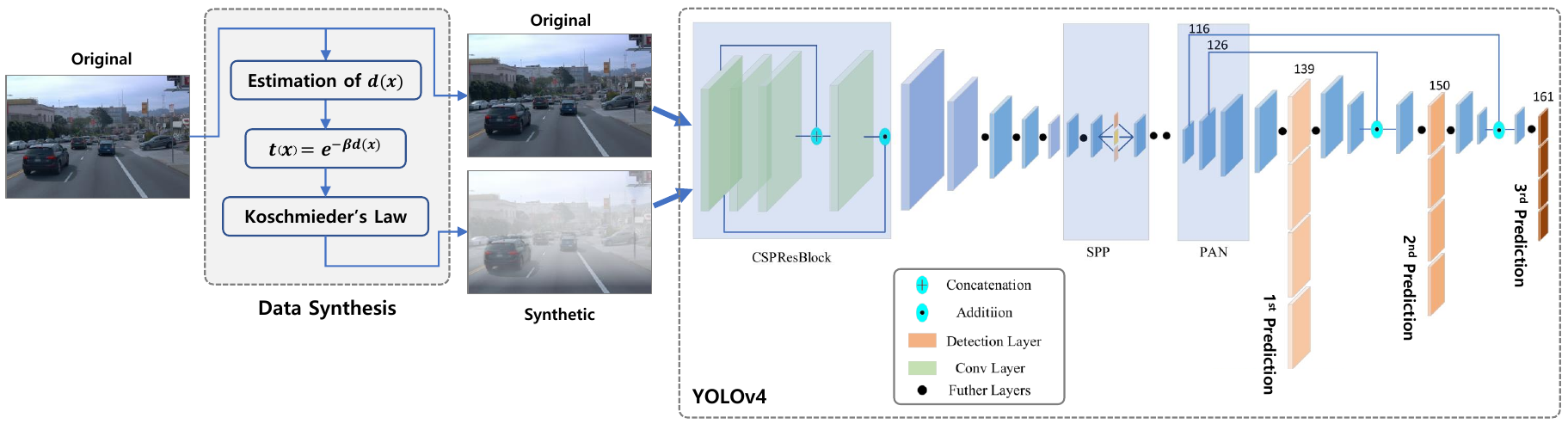}
\caption{Pipeline of the proposed method (the illustration of YOLOv4 is adapted from \cite{cai2021yolov4}).}
\label{fig2:proposedframework}
\end{figure*}

On the other hand, driving object detection has been one of the most well-studied topics in the current era of self-driving vehicles and autonomous robots. However, driving object detectors have a high probability of encountering domain shift because the environmental conditions may quickly change in practice. For instance, an object detector that was trained on clean-weather image data may not be able to perform well when facing inclement weather conditions like fog and haze. Fig. \ref{fig1:fogexamples}\textcolor{red}{c} illustrates typical real-world foggy scenes in driving scenarios where the visibility of the driver or autonomous system camera is substantially restricted. Therefore, improving the robustness of object detection models under such circumstances is extremely desired.

One effective approach to alleviating the problems that originated from domain shift is to increase the diversity of training data. In this paper, we aim to mitigate the effect of domain shift when the weather changes from clean to hazy, foggy, or dusty cases. Specifically, based upon the original clean image and Koschmieder’s law \cite{middleton1952vision}, we propose a data synthesis module that can generate high-quality hazy image data, as shown in Fig. \ref{fig1:fogexamples}\textcolor{red}{b}. This module can be implemented online or offline to train more robust and effective object detection models. By adopting YOLOv4 \cite{bochkovskiy2020yolov4} as the base object detector, we want to explore the effects on the object detection performance when applying the proposed data synthesis method in this paper.

The remainder of this paper is organized as follows: Section \ref{sec:relatedwork} briefly describes the relevant methods to this research. The proposed data synthesis module is proposed in Section \ref{sec:methodology}. The results of data synthesis and object detection are presented in Section \ref{sec:experiments}. Section \ref{sec:conclusions} concludes the paper.

\section{Related Works}
\label{sec:relatedwork}


\subsection{Haze Imaging Model}

Mathematically, a hazy image can be modeled as a per-pixel convex combination between the clean scene radiance and the global atmospheric light, this model is known as the haze imaging model or Koschmieder’s law \cite{middleton1952vision}:

\begin{equation}
    I(x)=J(x)t(x) + A(1-t(x)),
    \label{eq1:hazemodel}
\end{equation} 
where $x$ denotes pixel index, $I(x)$ represents the intensity of hazy image, $J(x)$ is the clean scene radiance, $A$ denotes the global atmospheric light, and $t(x)$ indicates the transmission map. When the global atmospheric light $A$ is homogenous, $t(x)$ can be expressed as \cite{tran2022novel}:

\begin{equation}
    t(x) = e^{-\beta d(x)},
    \label{eq2:transmission}
\end{equation} 
where $\beta$ represents the scattering coefficient of the atmosphere, and $d(x)$ denotes the depth information. From Eq. (\ref{eq1:hazemodel}), a hazy image $I(x)$ can be synthesized when the transmission $t(x)$ is given, and from Eq. (\ref{eq2:transmission}), the transmission is closely linked to the depth information $d(x)$ of the scene. Therefore, an accurate estimation of the scene depth information can result in a visually compelling synthetic hazy image. 

\begin{figure*}[t]
\centering
\includegraphics[width=1.0\textwidth]{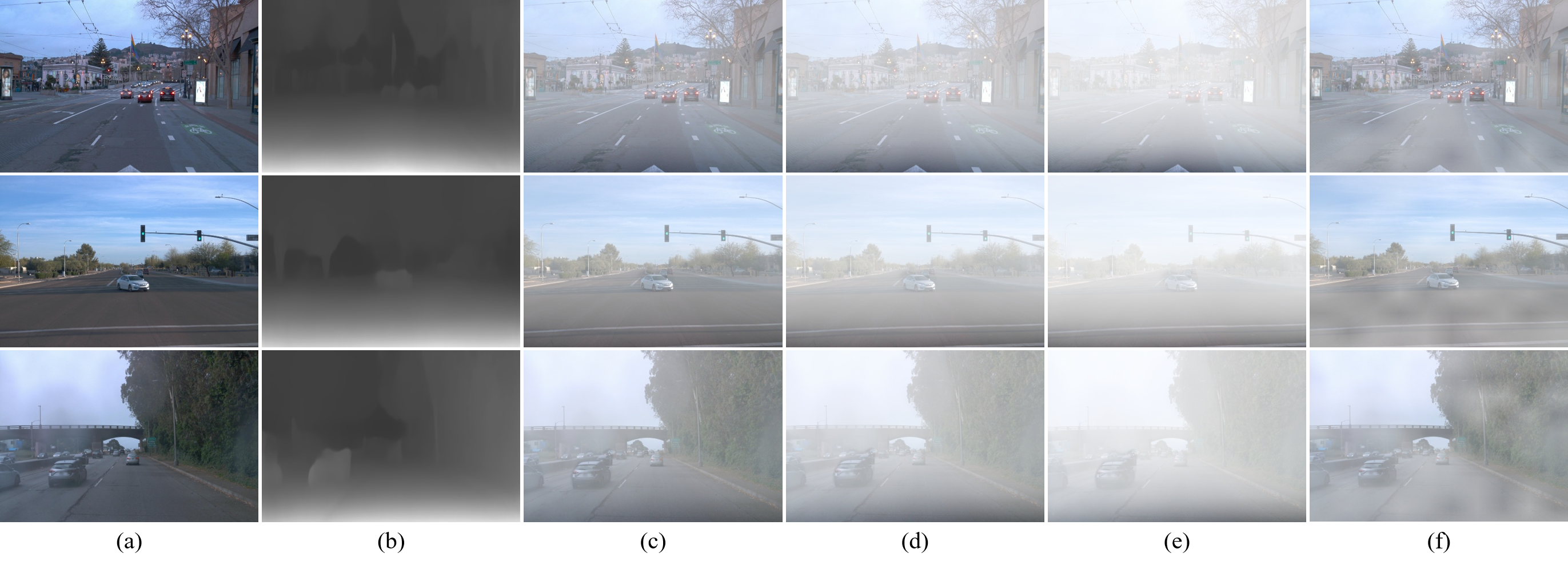}
\caption{Synthetic image data: (a) Original clean image, (b) Estimated depth, (c,d,e) Synthetic hazy images by the proposed method with different $\beta$ values: 1.0, 1.5, 2.0, respectively, note that larger value of $\beta$ results in denser haze, and (f) Synthetic hazy image based on transmission randomization.}
\label{fig3:synthetichaze}
\end{figure*}

\subsection{Depth Estimation Model}

The estimation of scene depth using only a single RGB input image is a challenging and long-standing problem in computer vision. Recently, Monodepth2 \cite{godard2019digging} is the method that has made a breakthrough in this field and has established a new baseline for single image depth estimation. Moreover, Monodepth2 has been validated on the KITTI benchmark \cite{geiger2013vision}, a driving scene dataset, which is closely related to the aim of this study. From those aspects, Monodepth2 is chosen as the depth estimation method in the proposed pipeline. In order to perform monocular depth estimation, the model has to estimate the egomotion between temporal image pairs during training. This process can be carried out by training a pose estimation network that takes as input a finite sequence of image frames and computes the camera transformations. Monodepth2 adopts a general encoder-decoder architecture as the backbone. Depth estimation and camera pose estimation are performed by two decoders. The encoder of Monodepth2 is a ResNet architecture \cite{he2016deep} which takes as input a single RGB image for the depth estimation. For the pose estimation, the encoder is constructed to receive a pair of image frames as input. Therefore, the first filter tensor is duplicated along the channel dimension to make the filter shape fit to the pose estimation encoder. On the other hand, the depth estimation decoder is constructed as a fully convolutional network with skip connections which allow the network to exploit the features in different resolution levels. A sigmoid activation function is utilized at the last layer in order to normalize the disparity map. In addition, the pose estimation decoder is also a fully convolutional network that predicts the rotation using an axis-angle representation.

\subsection{Object Detection Model}

Over the past few years, the object detection field has witnessed the dominant popularity of the one-stage object detector series, YOLO \cite{redmon2016you, redmon2017yolo9000, redmon2018yolov3, bochkovskiy2020yolov4}, which can achieve state-of-the-art performance on the MSCOCO benchmark \cite{lin2014microsoft}. The first YOLO model, YOLOv1 \cite{redmon2016you}, was proposed to combine the problems of drawing bounding boxes and recognizing class labels in a single end-to-end differentiable network, which had made a breakthrough in the field of object detection. One year later, YOLOv2 \cite{redmon2017yolo9000} made several improvements on top of YOLOv1 including the use of the batch normalization layer, higher resolution processing, and anchor box concept. After that, YOLOv3 \cite{redmon2018yolov3} was built upon previous models by designing further connections to the backbone network that can make predictions at three different levels of feature map resolution in order to improve the ability to detect small objects. So far, the most widely used version has been YOLOv4, since the follow-up versions have not made obvious progress. The architecture of YOLOv4 remains adequately uncomplicated so that it can be deployed to almost all types of hardware and applications. Therefore, we adopt YOLOv4 as the base object detector in this research. YOLOv4 utilizes CSPDarkNet53 \cite{bochkovskiy2020yolov4} as the backbone for feature extraction. For the neck part, PANet \cite{liu2018path} and SPP block \cite{he2015spatial} are adopted to increase the receptive field and separate out important features from the backbone. YOLOv4 employs the same head as YOLOv3 with the anchor-based detection strategy and three levels of detection granularity. During training, mosaic data augmentation \cite{bochkovskiy2020yolov4} is carried out which combines four cropped images together in order to teach the model to find smaller objects and pay less attention to surrounding scenes.

\begin{figure}[t]
\centering
\includegraphics[width=0.96\columnwidth]{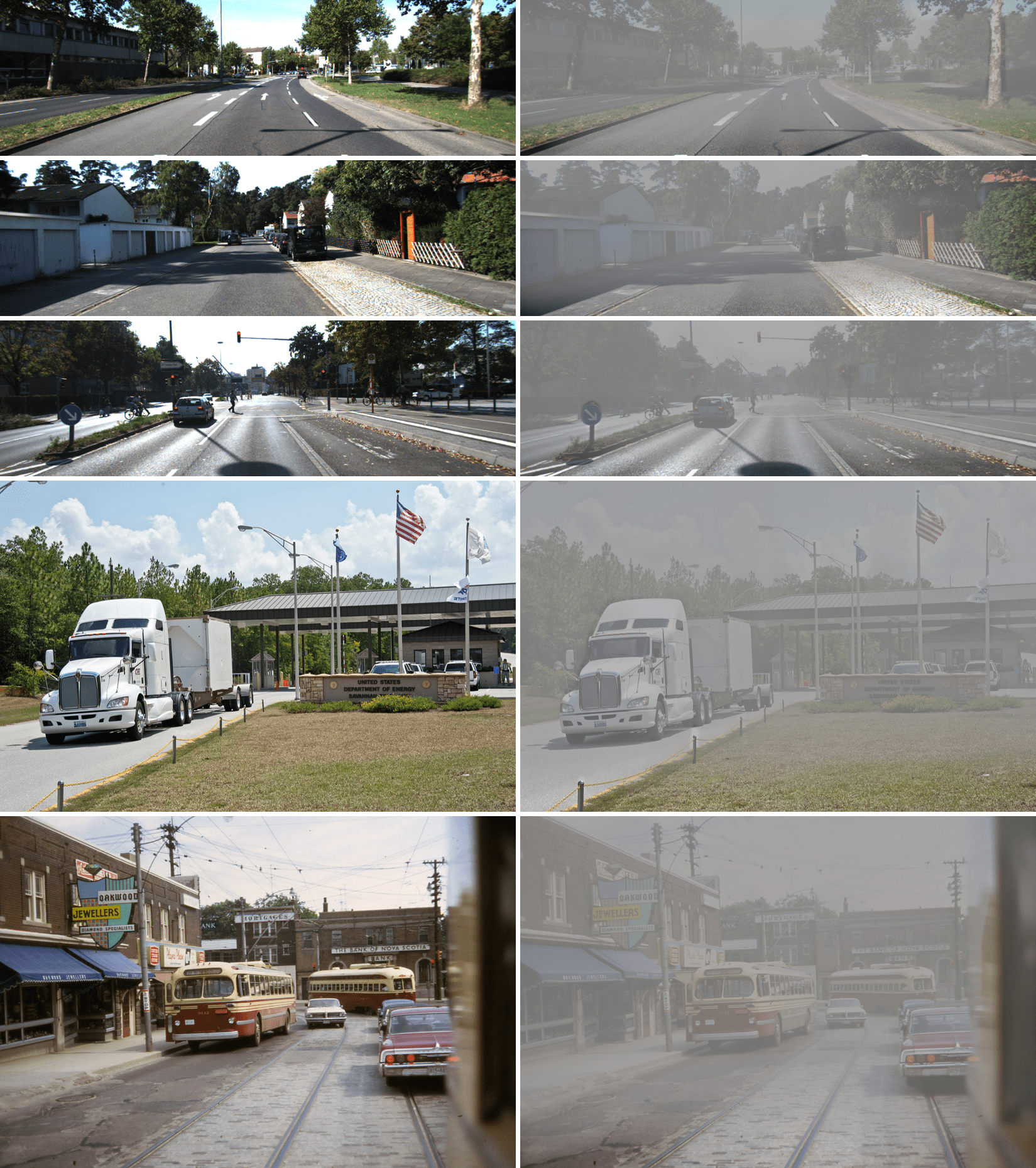}
\caption{Data synthesis results for KITTI (three top pairs) and MSCOCO image data (left: original, right: synthetic).}
\label{fig4:haze_synt}
\end{figure}

\begin{figure*}
  \centering
  
  \resizebox{0.98\textwidth}{!}{
  \begin{subfigure}{0.201\linewidth}
    \centering
    \includegraphics[width=1.02\linewidth]{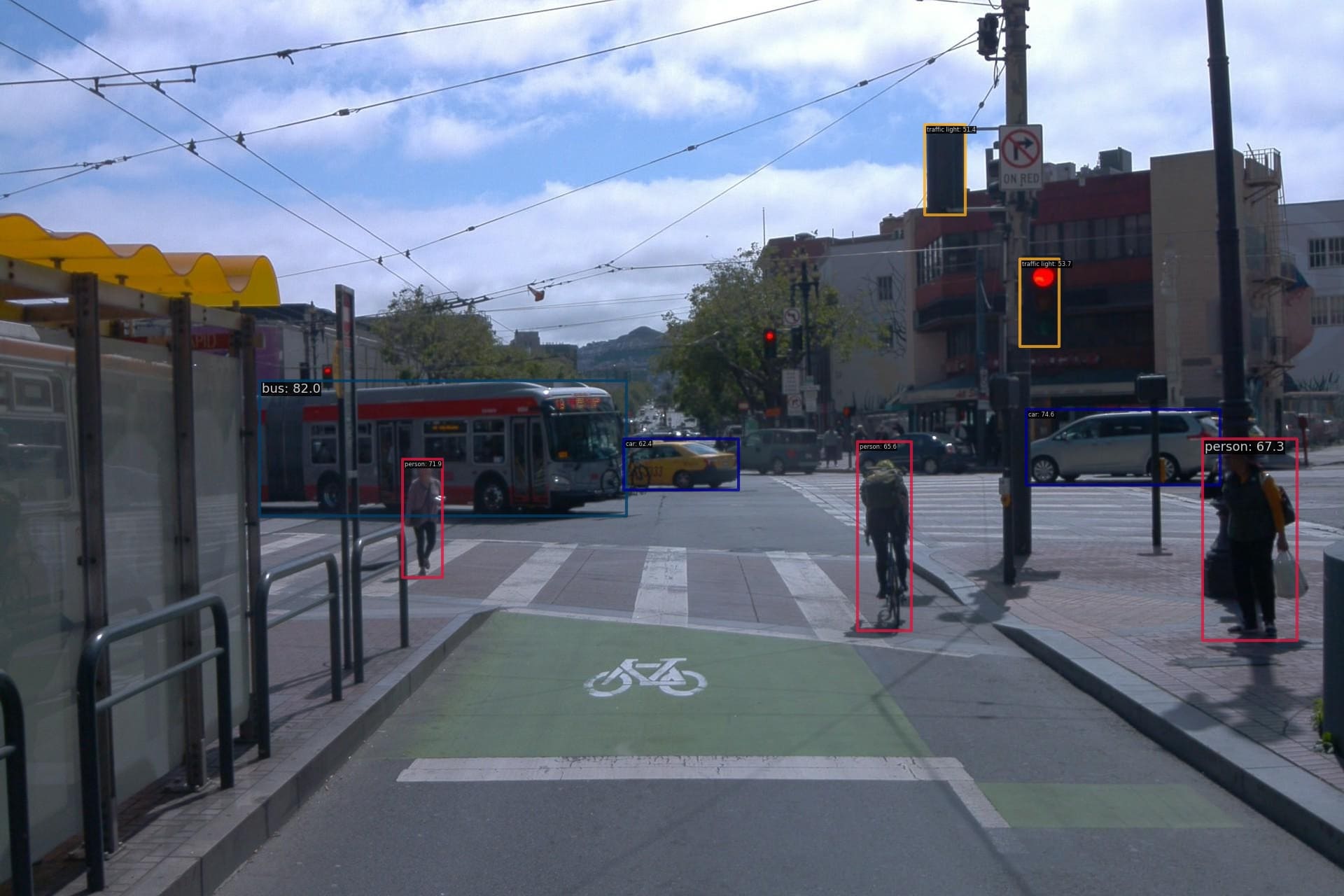}\\
    \vspace{0.05cm}
    \includegraphics[width=1.02\linewidth]{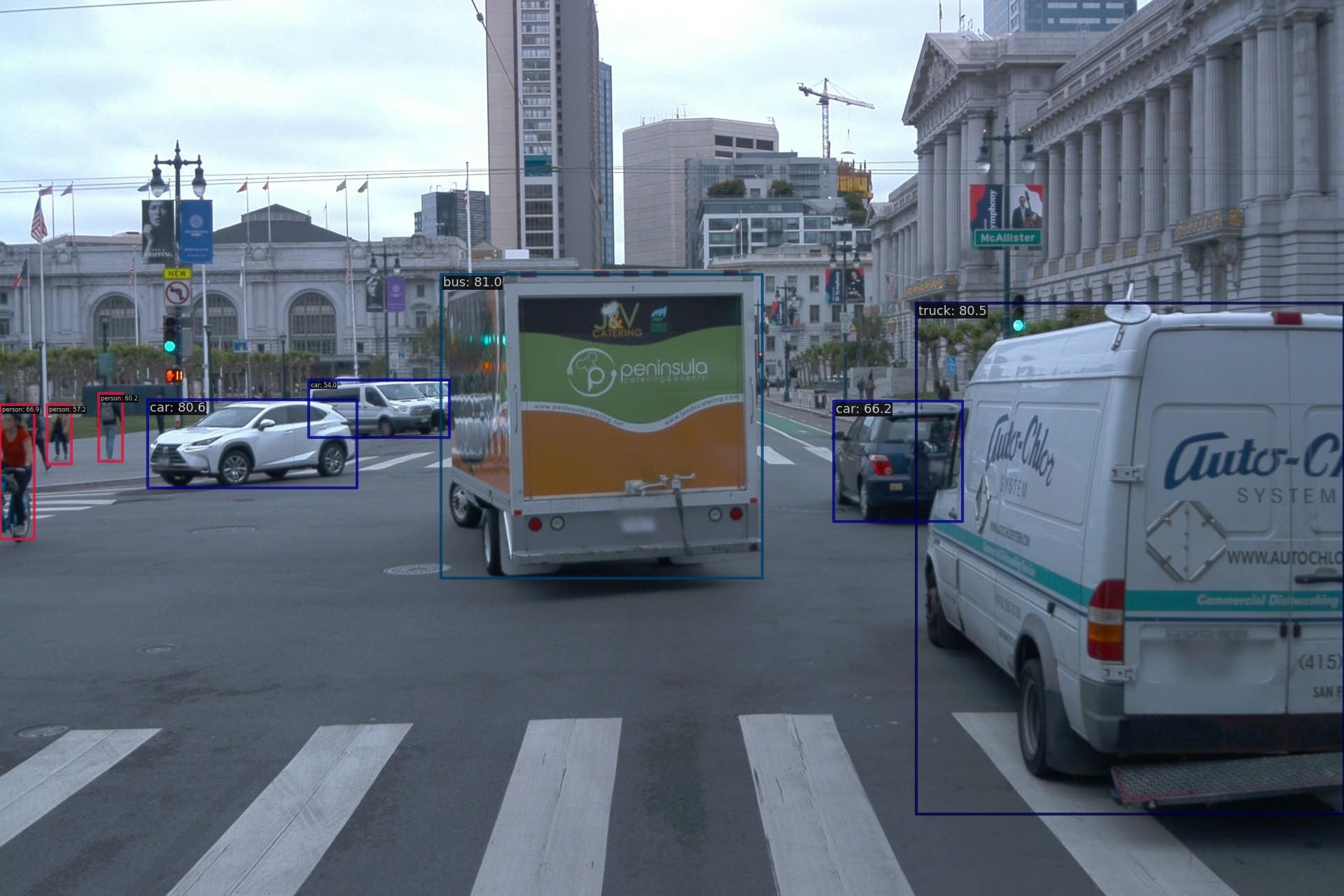}
    \captionsetup{font={footnotesize}}
    \caption{}
    \label{fig:waymo-det-a}
  \end{subfigure}
  \hfill
  \begin{subfigure}{0.201\linewidth}
    \centering
    \includegraphics[width=1.02\linewidth]{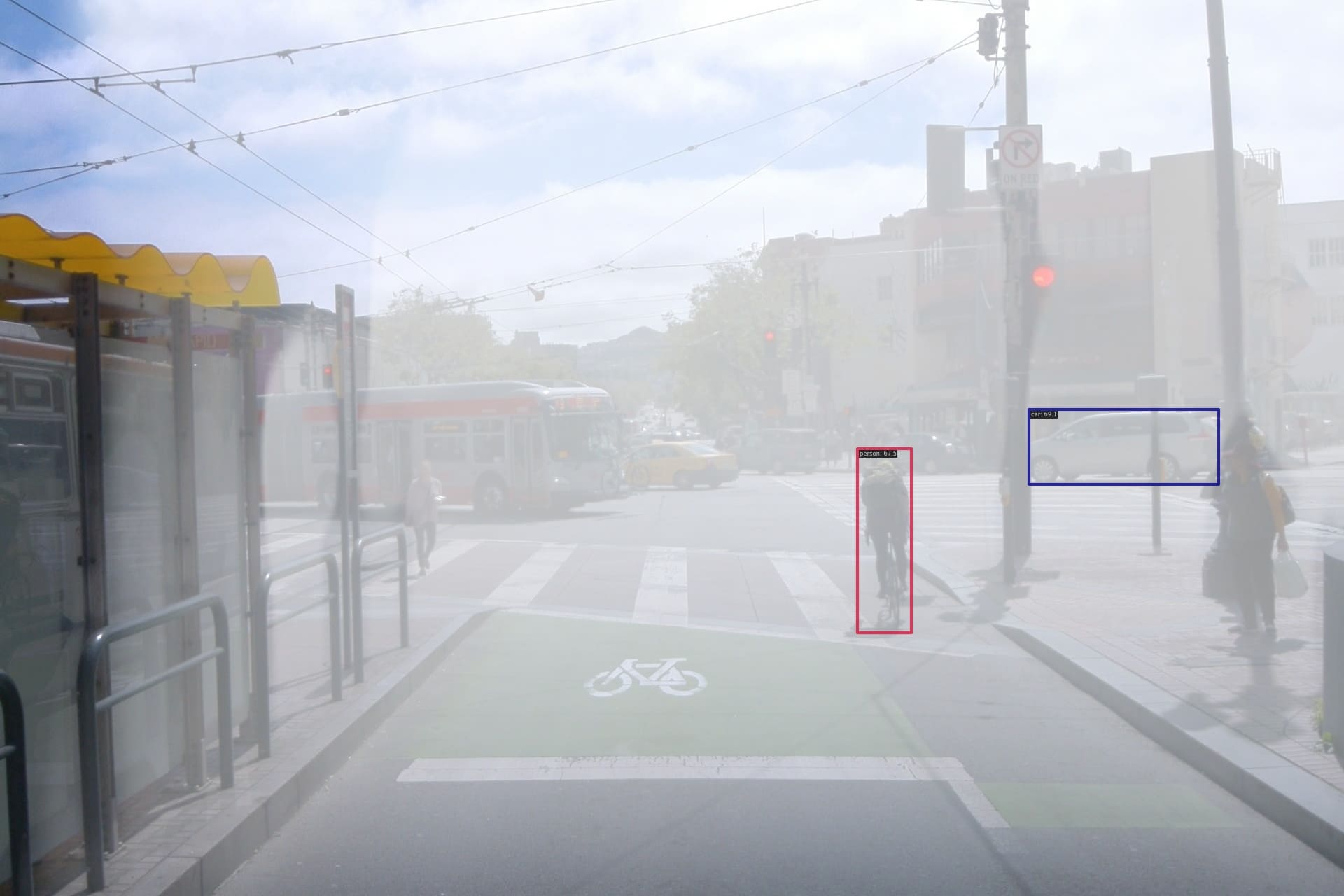}\\
    \vspace{0.05cm}
    \includegraphics[width=1.02\linewidth]{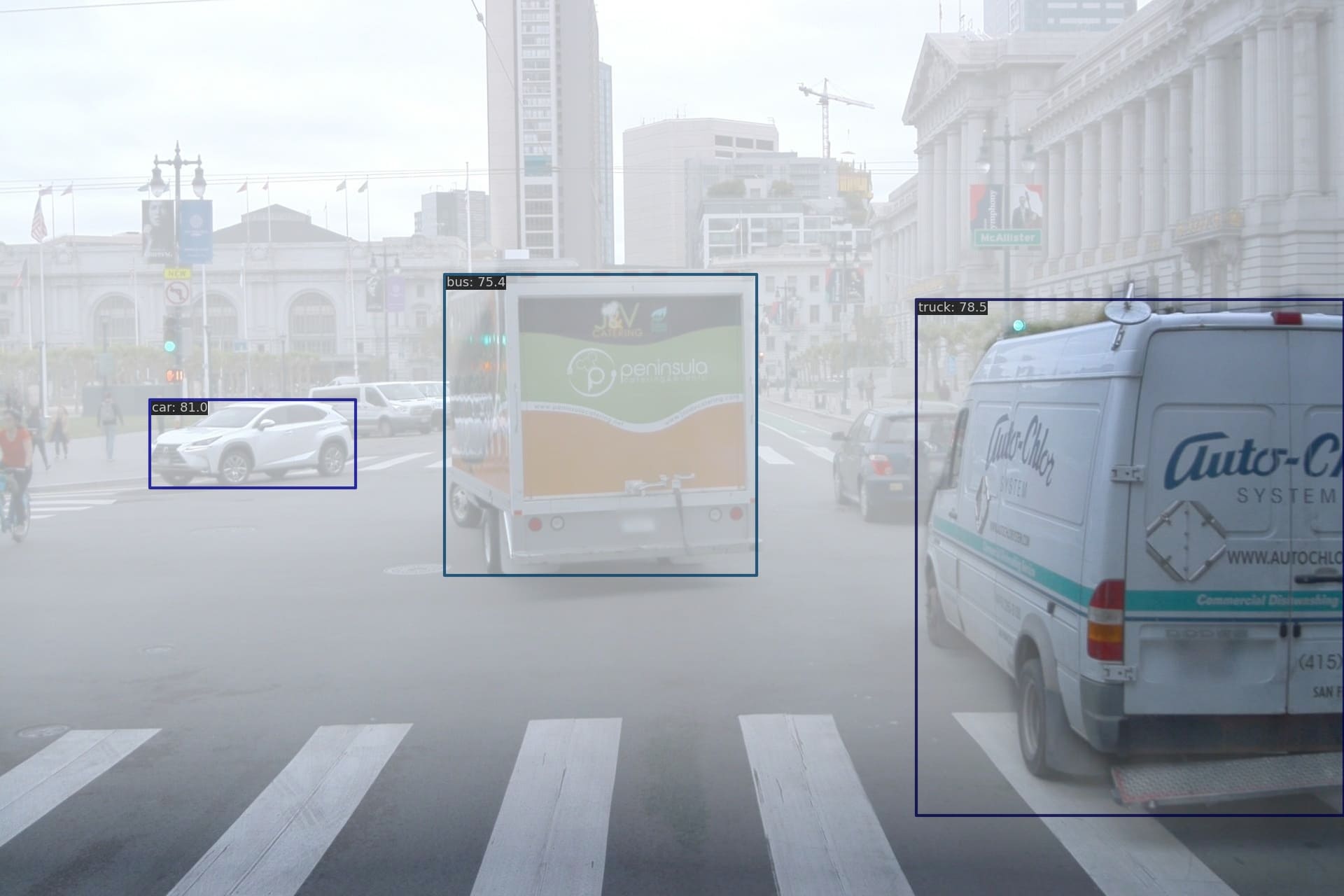}
    \captionsetup{font={footnotesize}}
    \caption{}
    \label{fig:waymo-det-b}
  \end{subfigure}
  \hfill
  \begin{subfigure}{0.201\linewidth}
    \centering
    \includegraphics[width=1.02\linewidth]{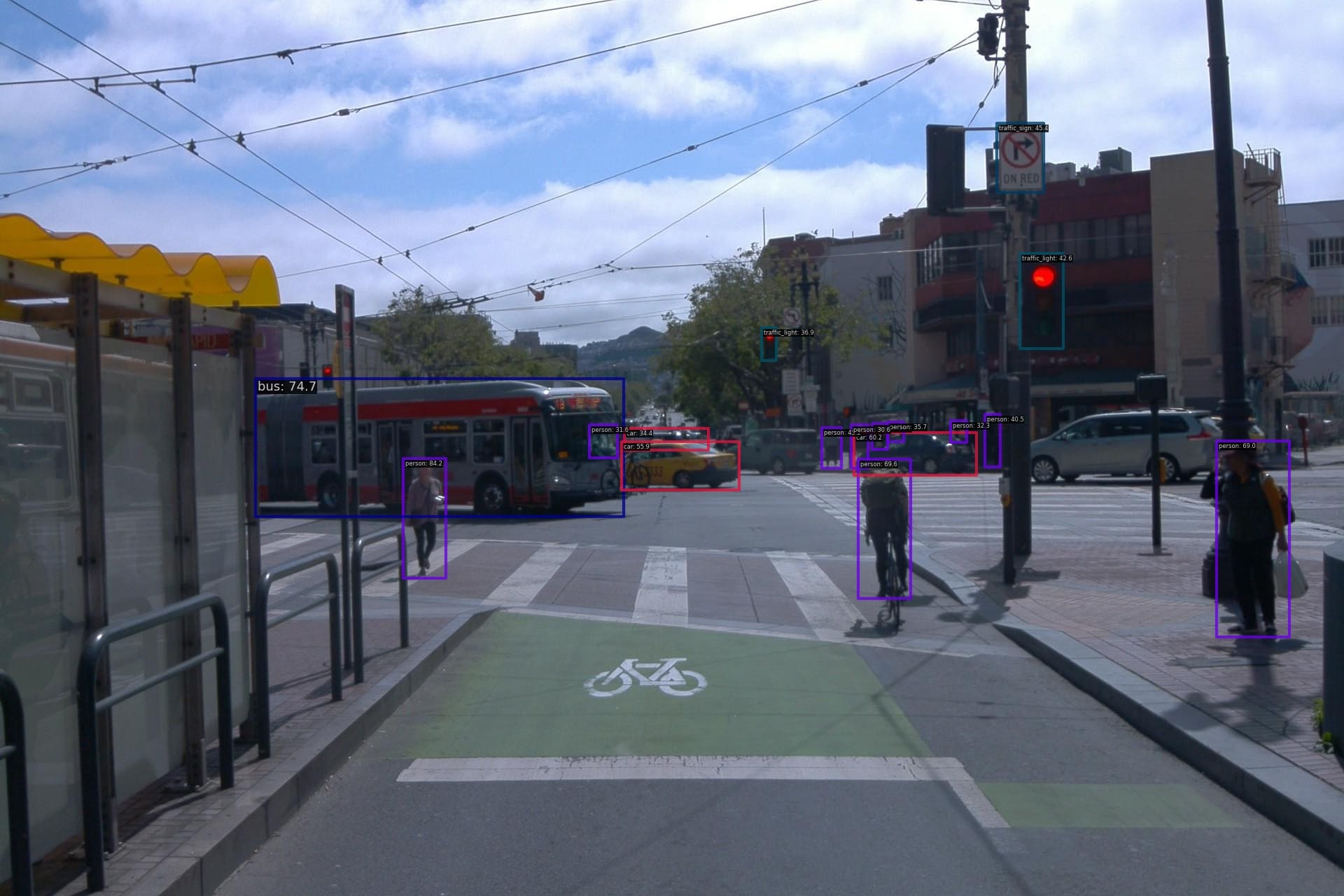}\\
    \vspace{0.05cm}
    \includegraphics[width=1.02\linewidth]{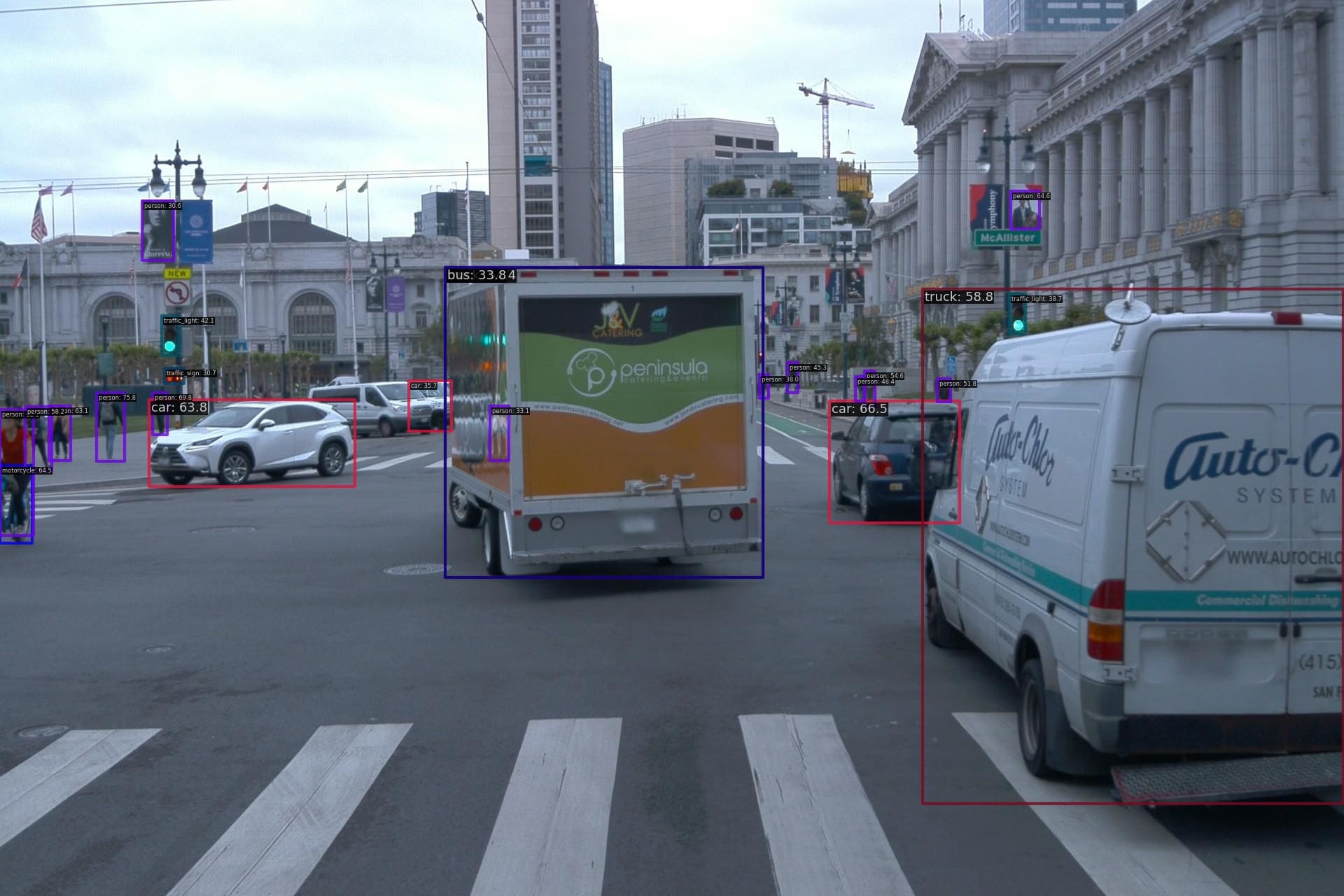}
    \captionsetup{font={footnotesize}}
    \caption{}
    \label{fig:waymo-det-c}
  \end{subfigure}
  \hfill
  \begin{subfigure}{0.201\linewidth}
    \centering
    \includegraphics[width=1.02\linewidth]{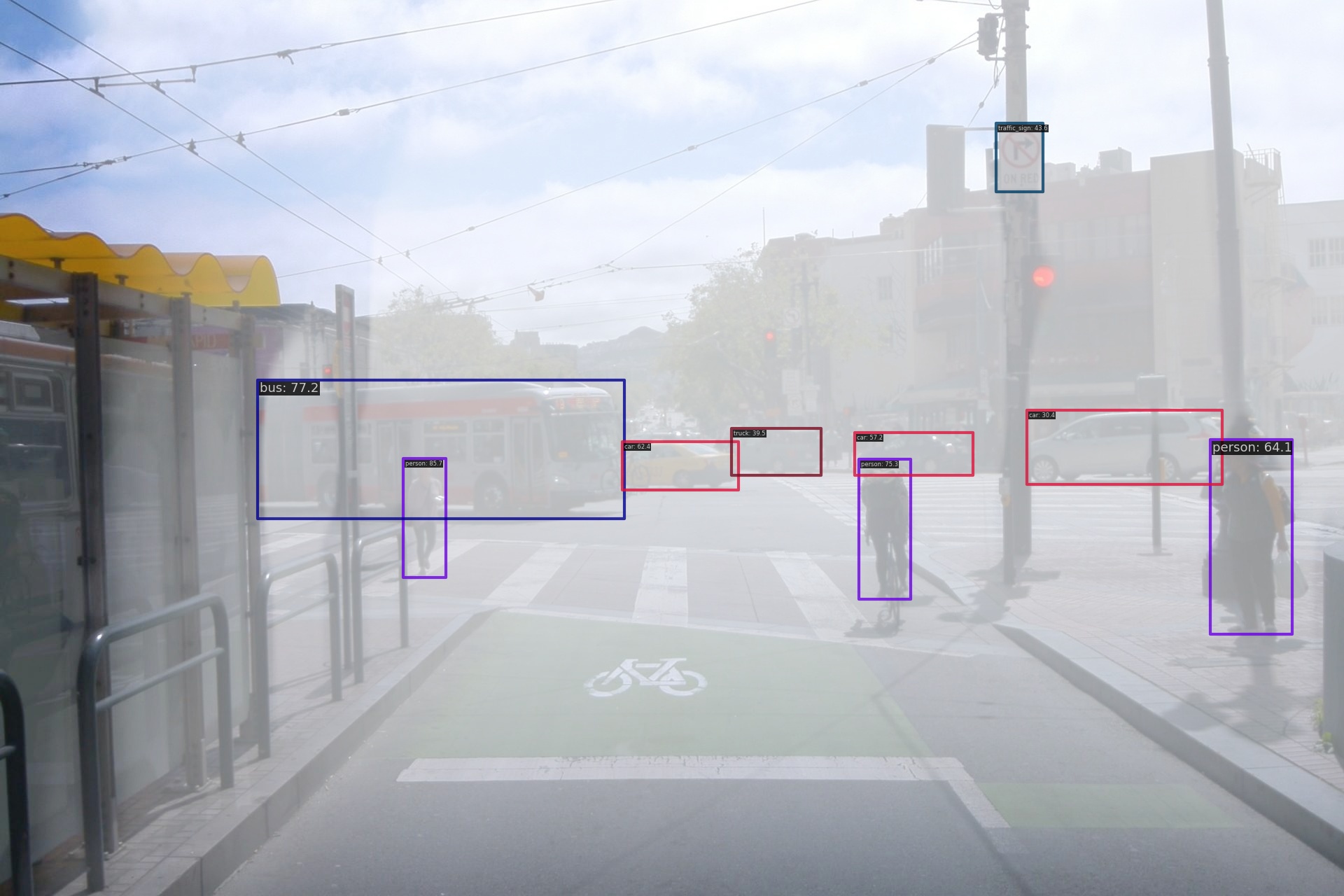}\\
    \vspace{0.05cm}
    \includegraphics[width=1.02\linewidth]{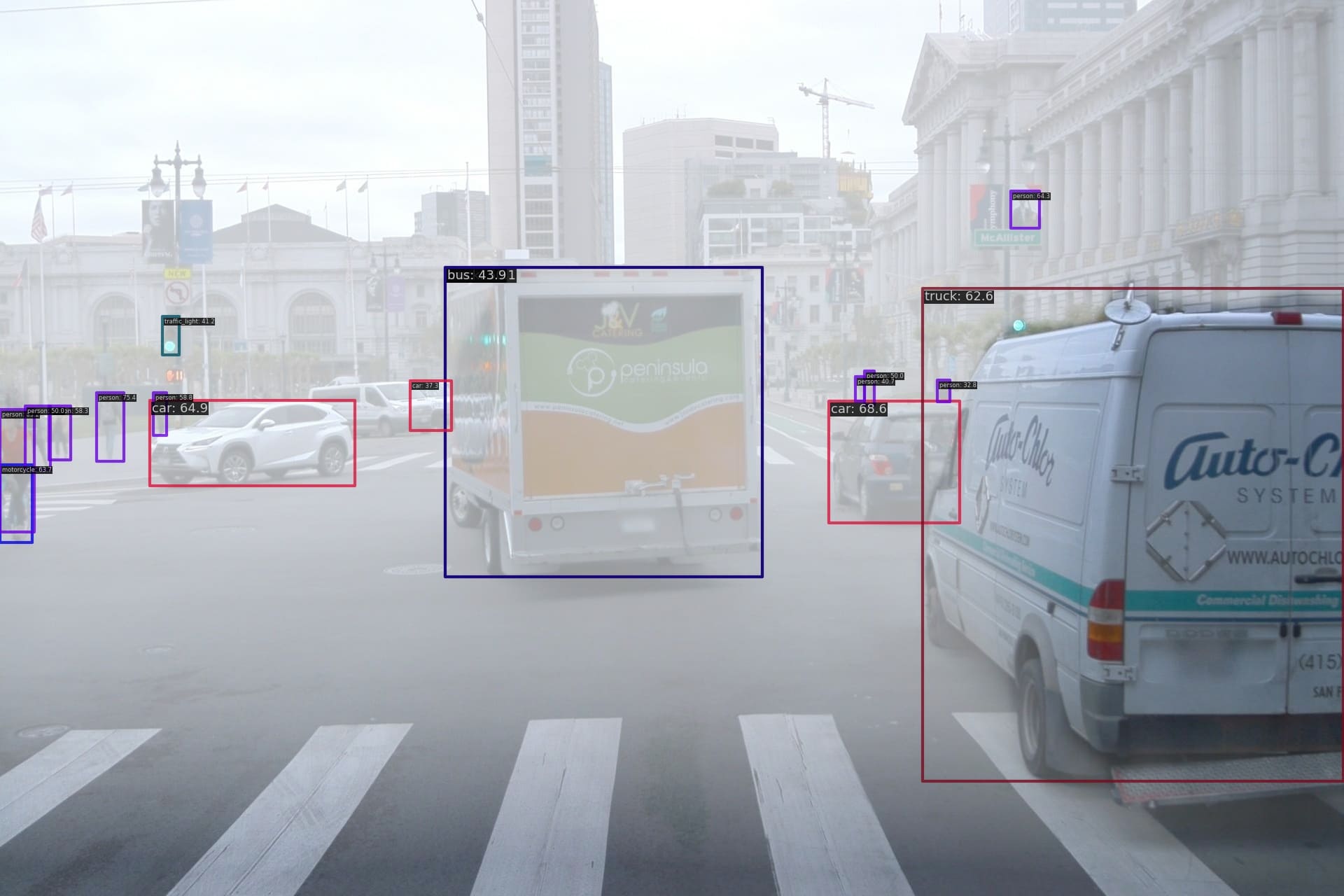}
    \captionsetup{font={footnotesize}}
    \caption{}
    \label{fig:waymo-det-d}
  \end{subfigure}}
  \caption{Object detection performance comparison: (a) YOLOv4 on clean scene, (b) YOLOv4 on hazy scene, (c) YOLOv4-Synt on clean scene, and (d) YOLOv4-Synt on hazy scene.}
  \label{fig4:waymo-det}
\end{figure*}

\begin{table*}
  \centering
  \caption{Quantitative comparison of object detection performance on different test sets.}
  \resizebox{0.8\textwidth}{!}{
  \begin{tabular}{cccccccccc}
    \toprule
      \multirow{2}{*}{\textbf{}} & \multicolumn{3}{c}{WAYMO} & \multicolumn{3}{c}{WAYMO-Haze} & \multicolumn{3}{c}{Foggy Driving}  \\
    \cmidrule{2-10}
      & mAP & Precision & Recall & mAP & Precision & Recall & mAP & Precision & Recall \\
    \midrule
     YOLOv4 & 41.58 & 0.5767  & \textbf{0.6360} & 36.17 & 0.5523 & 0.5457 & 72.54 & \textbf{0.3246} & 0.8182 \\
     YOLOv4-Synt (ours) & \textbf{42.65} & \textbf{0.5982} & 0.6347 & \textbf{41.91} & \textbf{0.5926} & \textbf{0.6228} & \textbf{77.88} & 0.3087 & \textbf{0.8774}   \\
    \bottomrule
  \end{tabular}}
  \label{tab1:comparisons}
\end{table*}

\section{Methodology}
\label{sec:methodology}

The proposed pipeline is illustrated in Fig. \ref{fig2:proposedframework}. As shown in Fig. \ref{fig2:proposedframework}, an input image is first passed through the proposed data synthesis module to generate a synthetic hazy image before being passed to the object detection model for training. The proposed data synthesis module includes three stages. In the first stage, depth map is estimated by Monodepth2 model. As mentioned throughout the paper, we aim at improving the robustness of object detectors in driving scenarios, hence, the used Monodepth2 model which was pre-trained on driving scene data \cite{geiger2013vision} can guarantee the quality of the extracted depth information and result in high-quality synthetic hazy images. In the second stage, transmission is computed by applying Eq. (\ref{eq2:transmission}). The value of scattering coefficient $\beta$ in Eq. (\ref{eq2:transmission}) is set to a randomly chosen real number in the range $[1.0, 3.0]$ in order to generate an arbitrary haze density for each input image and consequently the diversity of the synthetic data can be enhanced. In the last stage, the synthetic hazy image is produced by applying Koschmieder’s law as described in Eq. (\ref{eq1:hazemodel}), with $A$ randomly selected in the interval $[150, 255]$ to describe the airlight. Once the synthetic image data is obtained, it is combined with the original input to train the object detection model. The proposed module can be implemented in two fashions, offline and online, with similar performances. In the offline mode, the training data is synthesized prior to the training process, whereas in the online mode, the data is synthesized during training.

\begin{figure*}
  \centering
  
  \resizebox{0.98\textwidth}{!}{
  \begin{subfigure}{0.201\linewidth}
    \centering
    \includegraphics[width=1.02\linewidth]{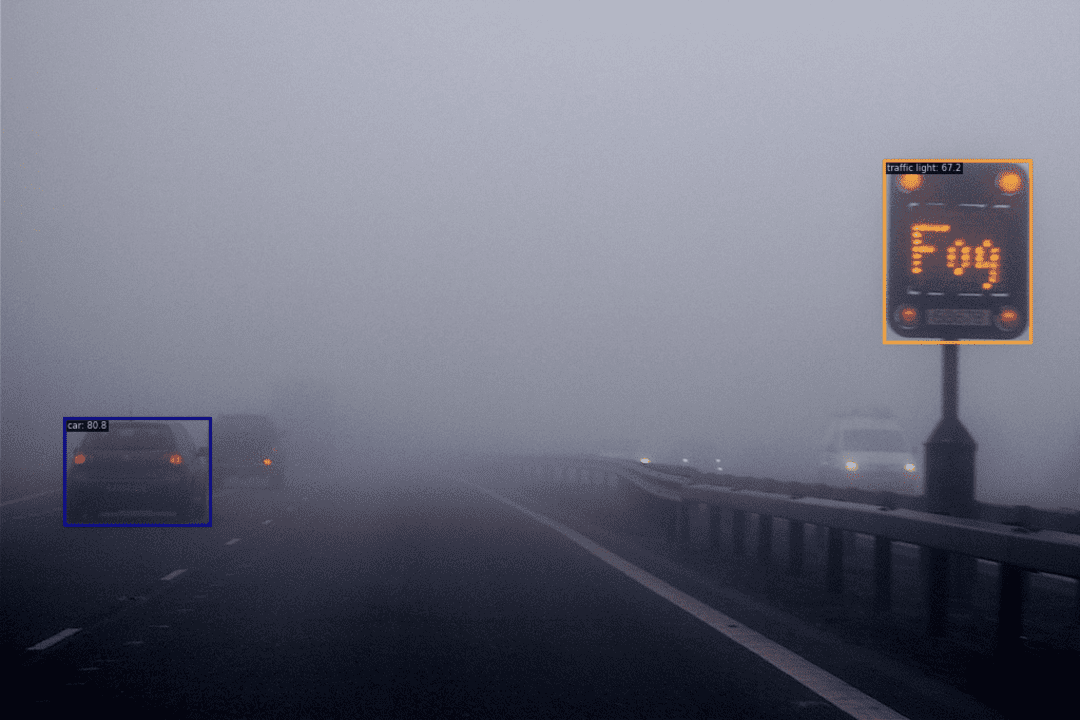}\\
    \vspace{0.05cm}
    \includegraphics[width=1.02\linewidth]{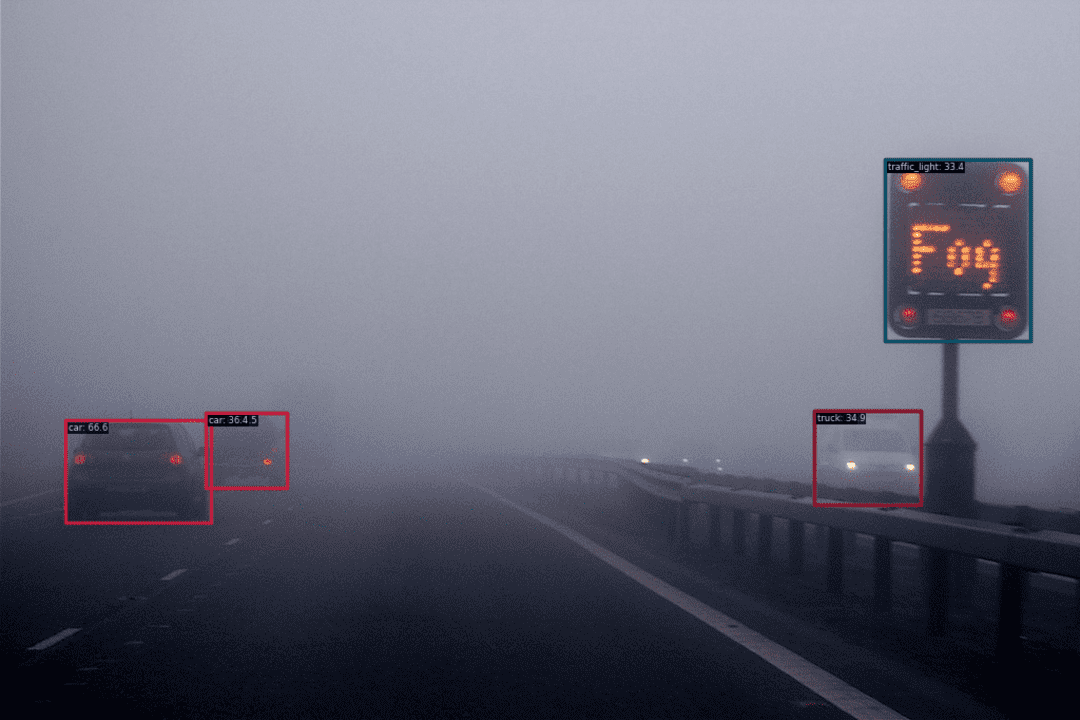}
    \label{fig:foggydriving-det-a}
  \end{subfigure}
  \hfill
  \begin{subfigure}{0.201\linewidth}
    \centering
    \includegraphics[width=1.02\linewidth]{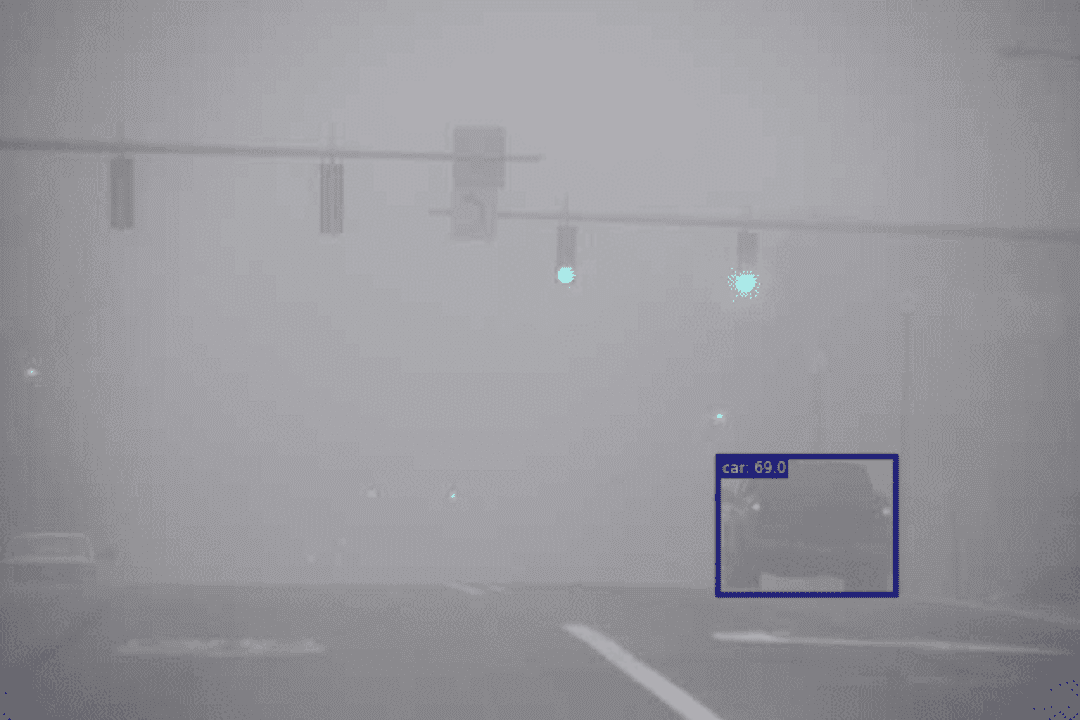}\\
    \vspace{0.05cm}
    \includegraphics[width=1.02\linewidth]{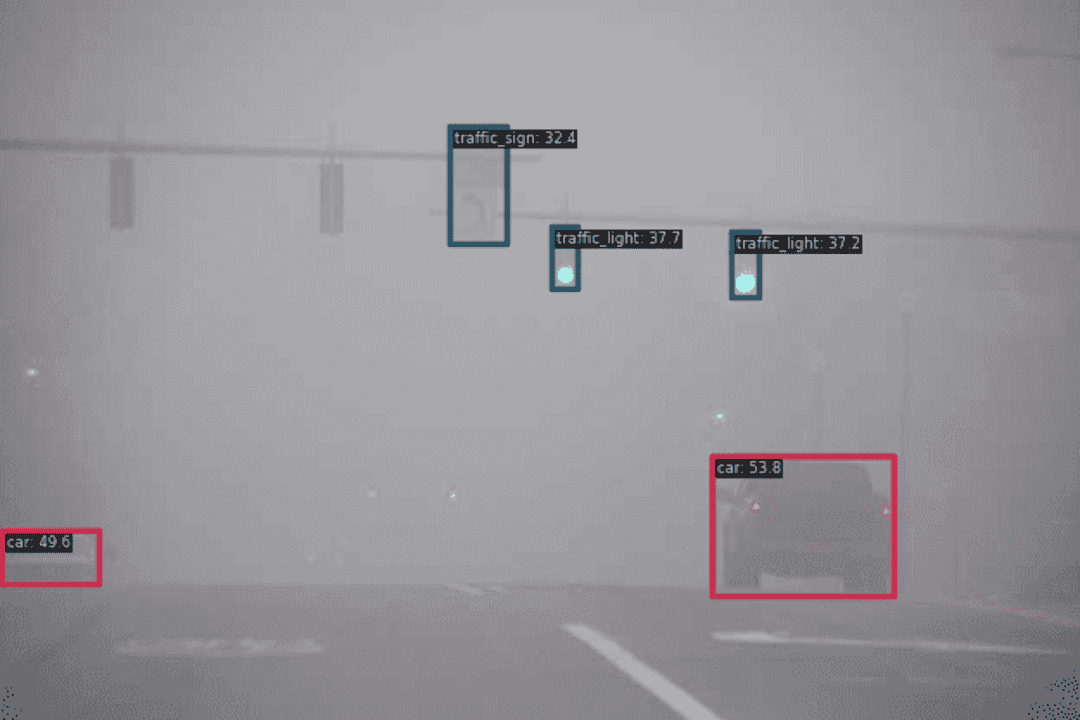}
    \label{fig:foggydriving-det-b}
  \end{subfigure}
  \hfill
  \begin{subfigure}{0.201\linewidth}
    \centering
    \includegraphics[width=1.02\linewidth]{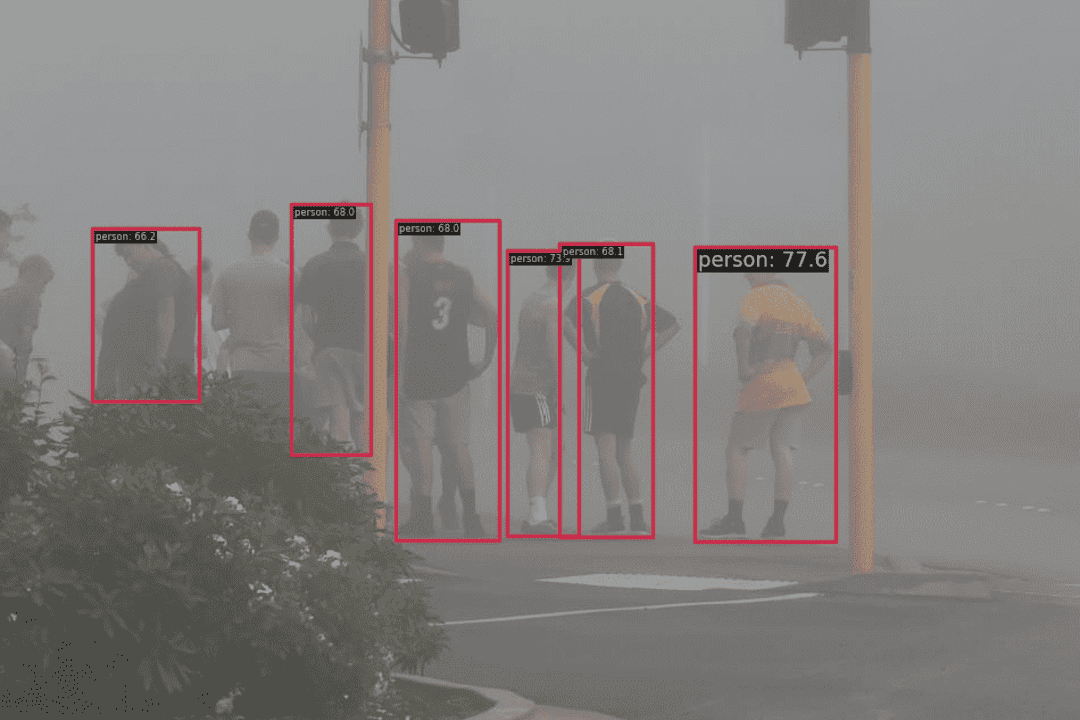}\\
    \vspace{0.05cm}
    \includegraphics[width=1.02\linewidth]{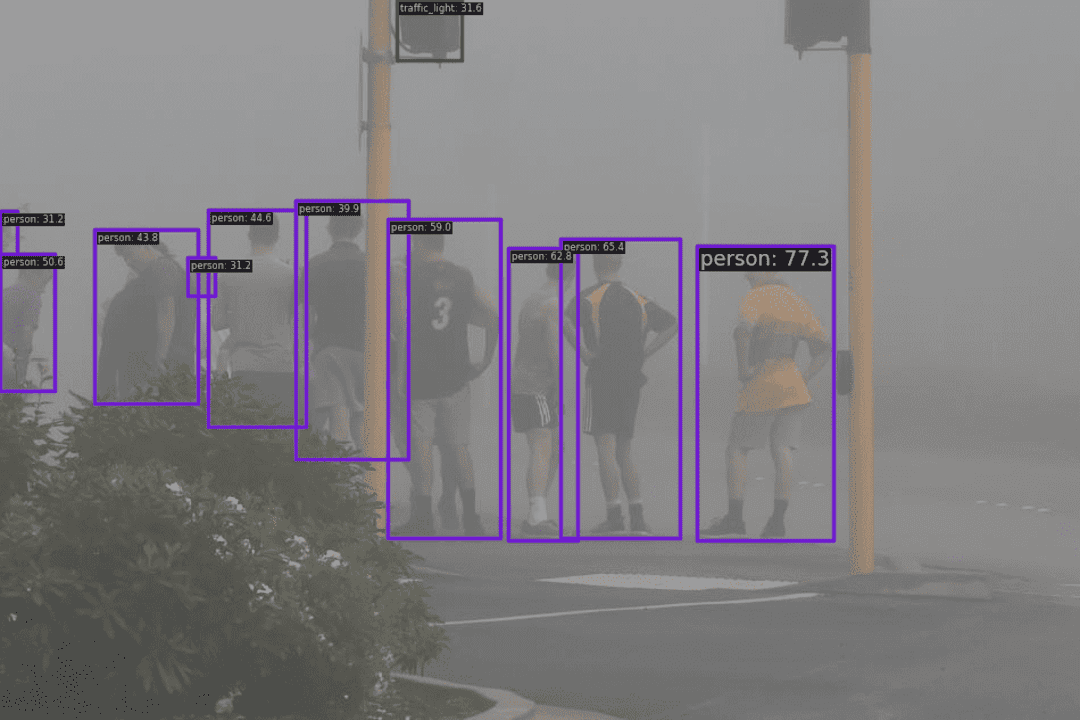}
    \label{fig:foggydriving-det-c}
  \end{subfigure}
  \hfill
  \begin{subfigure}{0.201\linewidth}
    \centering
    \includegraphics[width=1.02\linewidth]{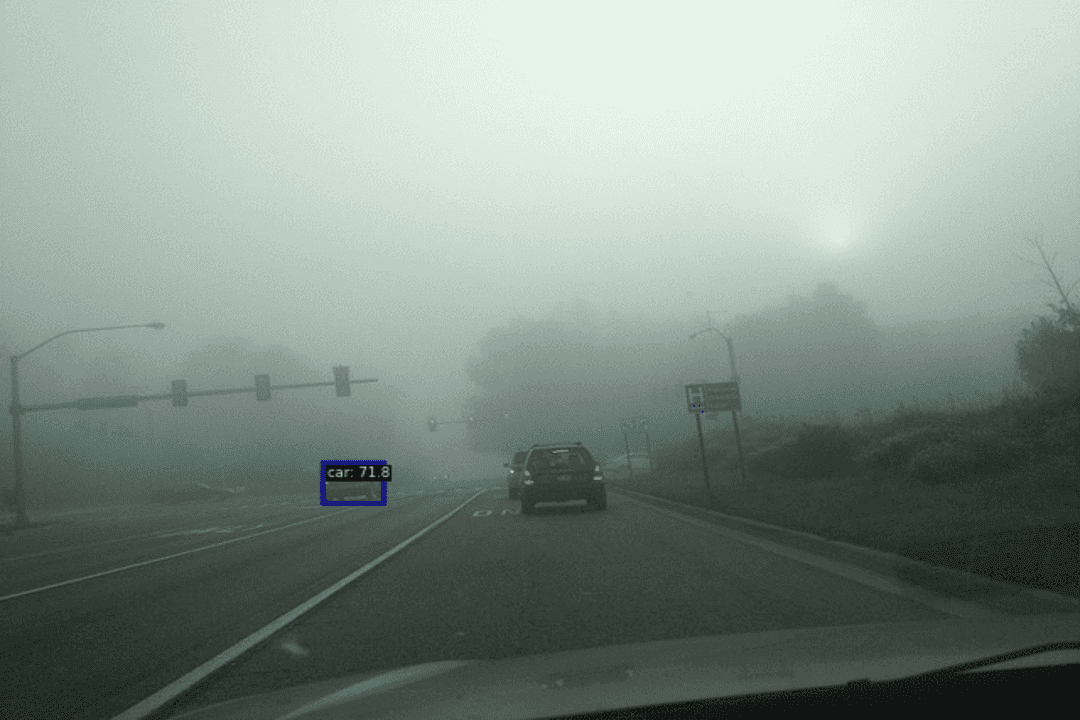}\\
    \vspace{0.05cm}
    \includegraphics[width=1.02\linewidth]{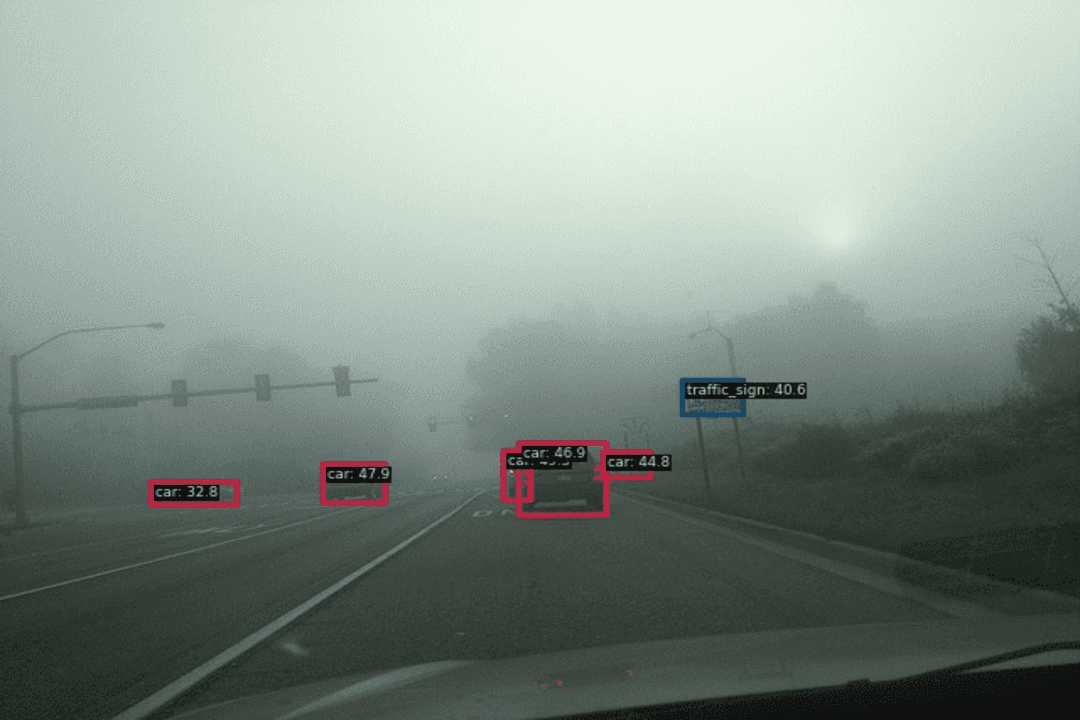}
    \label{fig:foggydriving-det-d}
  \end{subfigure}}
  \caption{Visual object detection results of YOLOv4 (top) and YOLOv4-Synt (bottom) on Foggy Driving dataset.}
  \label{fig5:yolodetonfoggydriving}
\end{figure*}

\section{Experiments}
\label{sec:experiments}


In this section, the visibility of the synthetic hazy data is first discussed, then the object detection performances on WAYMO \cite{sun2020scalability} and Foggy Driving \cite{sakaridis2018semantic} datasets are presented.

\subsection{Synthetic Image Data}

We first synthesize hazy image data from WAYMO dataset \cite{sun2020scalability}. We utilize a pre-trained Monodepth2 model which was trained on KITTI dataset to estimate the depth map of image data in WAYMO dataset because KITTI and WAYMO datasets contain similar image scenes. However, WAYMO dataset includes image data captured in five different views (front, front-left, side-left, front-right, side-right) from the driver's perspective while KITTI dataset contains the image data captured only from the front view. Hence, using all scenes in WAYMO dataset may not guarantee the production of adequate depth maps. Therefore, we select only front-view image data in WAYMO dataset, specifically, 10,000 front-view images for the training set and 100 front-view images for the test set. All selected images are captured during daytime period and in good lighting conditions. We term this synthesized dataset as WAYMO-Haze. Fig. \ref{fig3:synthetichaze} shows typical synthesized images using the proposed method with different settings of $\beta$ values in comparison with those of the method based on transmission randomization. As can be observed from Fig. \ref{fig3:synthetichaze}, the proposed method can produce more natural-looking hazy images. Further image synthesis results on KITTI and MSCOCO datasets are shown in Fig. \ref{fig4:haze_synt}.

\subsection{Object Detection Performance}

We compare the performance of YOLOv4 in two settings. In the first setting, we train YOLOv4 on only the original WAYMO dataset, and in the second setting, we train YOLOv4 on the original WAYMO dataset combined with the proposed data synthesis module. Note that the proposed module can be implemented in two modes: offline and online, as mentioned in Section \ref{sec:methodology}. Both modes can yield similar results. The online mode is performed in this experiment. That is, the data is synthesized during training. We term the model trained with the proposed synthesis module as YOLOv4-Synt.

The object detection performances on the WAYMO and WAYMO-Haze test sets are summarized in Table \ref{tab1:comparisons}. As the results shown in Table \ref{tab1:comparisons}, YOLOv4-Synt is slightly better than YOLOv4 on the original WAYMO test set. However, on the WAYMO-Haze test set, YOLOv4-Synt outperforms YOLOv4 with a significant gap. Fig. \ref{fig4:waymo-det} shows typical visual object detection results of YOLOv4 and YOLOv4-Synt on the WAYMO and WAYMO-Haze test sets. As shown in Fig. \ref{fig4:waymo-det}, YOLOv4 and YOLOv4-Synt produce relatively similar visual detection outcomes on the WAYMO test set, yet on the WAYMO-Haze test set, YOLOv4-Synt can clearly surpass YOLOv4 in terms of small and distant object detection.

In addition, object detection performance on natural hazy scenes from Foggy Driving dataset is also investigated. In this experiment, we directly utilize the pre-trained weights of YOLOv4 and YOLOv4-Synt from the preceding experiments to conduct a comparison on Foggy Driving dataset. The quantitative results are also given in Table \ref{tab1:comparisons} while typical visual detection results are shown in Fig. \ref{fig5:yolodetonfoggydriving}. As the quantitative results presented in Table \ref{tab1:comparisons}, the performance of YOLOv4-Synt is outstanding when compared to that of YOLOv4 in terms of mAP and Recall. Also, as can be seen in Fig. \ref{fig5:yolodetonfoggydriving}, YOLOv4-Synt can produce superior results against YOLOv4 and can help to obtain better localization and improve detection accuracy on distant and small objects.

\section{Conclusions}
\label{sec:conclusions}

This paper proposes a data synthesis module that can be implemented during the training process to improve the robustness of driving object detection models against domain shift. Domain shift usually results in a drastic degradation in the performance of deep neural network models. In order to deal with this problem, one of the most effective approaches is to upgrade the diversity of training data. To this end, an innovative data synthesis module is proposed which can be combined with the original dataset and implemented in the training process to train more robust and effective object detectors. The quantitative and qualitative results have shown that the proposed data augmentation approach can help to improve the robustness and effectiveness of object detection models on both in-domain and out-of-domain image data.

\balance
\bibliographystyle{IEEEtran}
\bibliography{ref.bib}

\end{document}